\documentclass[11pt]{article}

\usepackage[final]{acl}

\usepackage{times}
\usepackage{latexsym}
\usepackage{amsmath}
\usepackage{booktabs}
\usepackage{multirow}
\usepackage{multicol}
\usepackage{xcolor}
\usepackage[T1]{fontenc}

\usepackage[utf8]{inputenc}

\usepackage{microtype}

\usepackage{inconsolata}

\usepackage{graphicx}

\title{Lingua Custodia’s participation at the WMT 2025 Terminology shared task}

\author{Jingshu Liu \And Raheel Qader \And Gaëtan Caillaut \And Mariam Nakhlé  \\
 \hspace{-12.5cm}    Lingua Custodia, France \\
 \hspace{-12.5cm} \texttt{ \{jingshu.liu,mariam.nakhle,gaetan.caillaut,raheel.qader\}@linguacustodia.com}} 

\begin{document}
\maketitle
\begin{abstract}

 This paper presents Lingua Custodia’s submission to the WMT25 shared task on Terminology shared task targeting three language pairs: English-German, English-Spanish, and English-Russian.  Our approach builds on two recent instruction-tuned large language models \textit{Qwen3-4b} and \textit{Gemma3-4b-it} which we further finetune in a two-stage process. In the first stage, we perform supervised fine-tuning (SFT) on parallel data augmented with explicit terminology constraints. In the second stage, we apply Group Relative Policy Optimization (GRPO) with a custom reward function designed to encourage correct terminology usage while preserving general translation quality.  Experimental results on  English-German and English-Spanish language pairs show  $>=10$ points of accuracy gain in term match compared to the their baseline counterparts and $>=5$ points for English-Russian, while maintaining similar Comet-Kiwi scores. 
\end{abstract}

\section{Introduction}

Nowadays both classical encode-decoder  \citep{sutskever2014sequence, bahdanau2014neural, luong2015effective, vaswani2017attention} and decoder only LLM based neural machine translation systems\citep{alves2024toweropenmultilinguallarge, cheng2025seedx, qwen3mt2025}  have demonstrated strong performance across a wide range of translation tasks, with the latter increasingly favored due to their scalability, multilingual generalization, and ability to leverage instruction-tuning for domain adaptation. 

We leverage LLM-based MT systems due to their ease of fine-tuning and scalability, which make them well-suited for rapidly adapting to new domains and tasks. This paper describes our submission to the WMT25 Terminology translation task in English to German, English to Spanish and English to Russian.  The task aims to assess the extent to which machine translation models can utilize additional information regarding the translation of terminologies. 
Previous works on machine translation with terminology control can be grouped into two categories according to whether the method needs training the model with terminology information. One group incorporates the constraint during the inference time\citep{hokamp2017lexically, post2018fast, susanto2020lexically}.  An LLM trained without any terminology constraint can also be seen as this category with zero or few shot prompting during the inference.  These methods can typically satisfy most of the constraints but suffer from high computational cost and sometimes low translation quality because they always try to strictly apply the terminology constraint regardless of the correctness of the whole sentence.  The other group integrates lexical constraints during
training \citep{dinu2019, crego2016systran, song2019code, alves2024toweropenmultilinguallarge} by annotating terminology in order to guide the model to learn the enforcement of the translation constraints. The main disadvantage of these methods is the lack of guarantee of all constraints in the translations. Another limitation of these works is that they usually requires  extra resources such as a term dictionary, an annotator(a big LLM) to augment the data. 

Our work extends the prior approaches of \cite{ailem2021} and \cite{liu-etal-2023-lingua}, which follow the second line of methods. Furthermore, we build our systems by finetuning two of the most prominent open large language models: \textit{Qwen3} \citep{qwen3mt2025} and \textit{Gemma3} \citep{gemmateam2025gemma3}.

We evaluate our work on the WMT25 terminology track1 testsets. Since the reference is not released by the time of writing this paper, we evaluate our system by a simple naive strict match with respect to the target constraint completed by the Comet-Kiwi \citep{rei-etal-2022-cometkiwi} score to make sure that the translation ability is sustained. Results show that our proposed approach outperforms the baseline by an average of more than 10 points in term match  accuracy.

\section{Method}

In this section we present our systems for the terminology task. We focus on the sentence-level track and our all of our proposed systems are multilingual. Therefore one system is applied to all the language pairs during the inference time. 

\subsection{Background}

In order to annotate the Enligsh-German and English-Spanish data, we choose to exploit the dictionary extracted by  \cite{liu-etal-2023-lingua} which builds a large bilingual terminology in a unsupervised manner. Alternative approaches such as \citet{HazemMorin2016} and \citet{Liu2018} can also achieve the same goal bu they require heavy computation, while \citet{Artetxe_2016}  learns only single word bilingual lexicon.   We follow the same extraction setting as in \citet{liu-etal-2023-lingua} to keep the process efficient: limit the candidate ngams to 5.  

Our systems are finetuned models from two popular  instruction LLMs Qwen3-4b \citep{yang2025qwen3} and Gemma3-4b-it \citep{gemmateam2025gemma3}. Both models stand out as high-performing models with open weights and efficient inference capabilities.  Besides, they have been optimized for instruction following and multilingual understanding, which fit perfectly  our scenario. With 4 billion parameters, they are well-suited for fast fine-tuning and efficient inference in constrained environments, enabling manageable  deployment in our production environment without the extensive resource demands of larger models.

\subsection{Instruction data generation}

We use public open general bilingual data to finetune the LLMs. For English-German and English-Spanish pairs, we use \textit{Common Crawl} \citep{commoncrawl2023} as our database. As for the English-Russian pairs, we take the train dataset provided by WMT25 which is a collection of various open data sources \footnote{ https://www2.statmt.org/wmt25/mtdata/mtdata.recipes.wmt25-constrained.yml}.  We will explain the details of the data cleaning and filtering in the Section 3.1. 

To annotate the instruction data with terminology control, we first employed \textit{Meta-Llama-3.1-405B-Instruct} \citep{grattafiori2024llama3herdmodels} to generate 10 diverse instruction templates. For example: \textit{Translate the following sentence from [src\_lang] to [tgt\_lang] while adhering to the specified terminology: [mapping\_list] should be used as a reference.} This variation is intended to improve model generalization across diverse instructions. The \textit{[mapping\_list]} is obtained by matching the sentence against our bilingual terminology dictionary, extracted using the method described in Section 2.1. To construct the final training set with terminology control, each parallel sentence is parsed using the dictionary, one instruction template is randomly selected, and the result is transformed into either the  Qwen3 or Gemma3 chat format. Following the official recommendations of the Qwen3 authors, we disable the \textit{thinking} mode by inserting empty content between the \textit{thinking} tags and exclude the corresponding thinking tag tokens from the training loss computation.

\subsection{Model finetuning}

Our finetuning process consists of two main phases:

\begin{itemize}
    \item \textbf{Supervised fine-tuning (SFT)}
    \item \textbf{Group Relative Policy Optimization (GRPO)}
\end{itemize}

 In the first phase, we perform SFT on our curated multilingual dataset, enabling the model to learn the desired instruction-following and terminology control behavior. In the second phase, we apply GRPO to further align the model’s outputs with task-specific quality objectives, enhancing adherence to the provided terminology constraints while keeping the translation quality.
  
For translation quality, we define a \textbf{BLEU-based reward}:
\[
R_{\text{BLEU}}(y, y^\ast) = \frac{\text{BLEU}(y, y^\ast)}{100}
\]
where $\text{BLEU}(y, y^\ast)$ is the sentence-level BLEU score computed using \texttt{SacreBLEU} \citep{post-2018-call}. The score is normalized to the range $[0, 1]$ to scale the reward.  

For terminology adherence, we design a \textbf{constraint-following reward} that parses the user instruction for terminology mappings of the form ``source term $\rightarrow$ target term''. This reward computes the proportion of target terms that appear in the generated translation:
\[
R_{\text{term}}(y, M) = \frac{|\{t \in M_{\text{target}} : t \subset y\}|}{|M|}
\]
where $M$ is the set of source--target terminology pairs extracted from the prompt, and $M_{\text{target}}$ is the set of target terms. If no terminology is specified in the prompt, we assign a default reward of $1.0$.  

These rewards are computed per sample and can be combined in a weighted manner during GRPO training to simultaneously improve general translation quality and enforce terminology constraints.

\section{Experiments}

 Our experiments are designed to compare the LLM baselines with our fine-tuned Qwen3 and Gemma3 models, measuring both translation quality performance and terminology adherence on the WMT25 benchmarks. The following sections detail the training datasets, the experimental settings, and the results obtained.

\subsection{Data}

We apply \textit{LaBSE} \citep{feng-etal-2022-labse} on the \textit{Common Crawl} and WMT25 English-Russian collected datasets. Pairs with a similarity score below 0.9 are discarded. From the remaining data, we randomly sample 10,000 parallel sentences for each language pair to construct the supervised fine-tuning dataset. For the GRPO training phase, we sample an additional 1,000 parallel sentences per language pair from the filtered data. Hence our final training dataset contains 33,000 samples.  The dataset is then automatically annotated with terminology control information with our unsupervised dictionaries. 

An example of a training sample:

\begin{figure}[h!]
\centering
\footnotesize
\begin{minipage}{0.95\linewidth}
\textbf{User prompt:} 
\begin{verbatim}
Generate a Spanish translation that accurately 
reflects the terminology specified in 
clause → cláusula.

Text to translate: To this end, it must comply 
with the WTO requirements and, in particular, 
with the GATT enabling clause cláusula of 1979.
\end{verbatim}

\textbf{Assistant response:}
{\color{blue}\texttt{ A tal fin, debe cumplir con los requisitos de  la OMC y, en especial, con la cláusula de habilitación del GATT de 1979.}}
\end{minipage}
\caption{Example of a training sample with terminology control. Only the bleu part is calculated for the loss during the trainig. We vary the user prompt using the 10 templates previously generated.}
\label{fig:training-sample}
\end{figure}

\subsection{Settings}

For the supervised finetuning (SFT) phase, we train Qwen3-4B and Gemma3-4B-IT using the AdamW optimizer with a learning rate of $5\times10^{-6}$,  weight decay of 0.01, and 1000 warmup steps. Each training batch per device contains 16 samples, and the maximum sequence length is set to 2048 tokens. Training is performed in \texttt{bf16} mixed-precision mode for computational efficiency. Gradient clipping is applied at a maximum norm of 1.0. The finetuning runs on 4~H100~GPUs using FSDP2 \citep{zhang2024simplefsdpsimplerfullysharded} for memory efficiency.  

For the GRPO phase, the per-device batch size is reduced to 8, and for each sample 16 distinct completions are generated to compute rewards. The reward function combines a BLEU-based score and a terminology adherence score with equal weighting (see Section2.3). GRPO training is carried out on 4 nodes of 4 H100 GPU machines using DeepSpeed ZeRO-3 \citep{rajbhandari2020deepspeedzero3}, \texttt{bf16} precision, and the AdamW optimizer with the same learning rate and weight decay as in SFT.

\subsection{Results}

We evaluate our systems on the translation constraint success rate by a simple strict match because by the time of our naive evaluation, the reference was not available. We report our results obtained with baseline, SFT finetuned and SFT+GRPO models on the WMT25 testset with the these settings of WMT25:

\begin{itemize}
    \item \textbf{No terminology}: the system is only provided with input sentences/documents.
    \item \textbf{Proper terminology}: the system is provided with input texts and terminology dictionaries.
    \item \textbf{Random terminology}: the system is provided with input texts and translation dictionaries that contain words randomly drawn from input texts. 
\end{itemize}

\begin{table*}[ht]
\centering
\setlength{\tabcolsep}{3pt}
\begin{tabular}{l l | cc|cc|cc}
\toprule
\multirow{2}{*}{Model} & \multirow{2}{*}{Setting} 
& \multicolumn{2}{c|}{EN$\rightarrow$DE} 
& \multicolumn{2}{c|}{EN$\rightarrow$ES} 
& \multicolumn{2}{c}{EN$\rightarrow$RU} \\
& & comet-kiwi & term\% & comet-kiwi & term\% & comet-kiwi & term\% \\
\midrule
\multirow{3}{*}{Qwen3-4b} 
  & proper & \textbf{0.7979} & 0.5525 & 0.8106 & 0.5242 & 0.8035 & 0.4280 \\
  & random & \textbf{0.7917} & 0.7724 & 0.8081 & 0.8202 & \textbf{0.7959} & 0.7228 \\
  & noterm & 0.8022 &   -    & 0.8183 &   -    & 0.8064 &   -    \\
\midrule
\multirow{3}{*}{Gemma3-4b-it} 
  & proper & 0.7887 & 0.5562 & 0.8106 & 0.5465 & 0.8025 & 0.4008 \\
  & random & 0.7831 & 0.7659 & 0.8005 & 0.8063 & 0.7893 & 0.7120 \\
  & noterm & 0.8000 &   -    & 0.8219 &   -    & 0.8094 &   -    \\
\midrule
\multirow{3}{*}{Qwen3\_SFT} 
  & proper & 0.7755 & 0.5543 & 0.8060 & 0.5892 & 0.7942 & 0.4144 \\
  & random & 0.7741 & 0.8065 & 0.8035 & 0.8709 & 0.7872 & 0.7246 \\
  & noterm & 0.7901 &   -    & 0.8147 &   -    & 0.8030 &   -    \\
\midrule
\multirow{3}{*}{Gemma3\_SFT} 
  & proper & 0.7924 & 0.5561 & \textbf{0.8148} & 0.5725 & \textbf{0.8038} & 0.3988 \\
  & random & 0.7893 & 0.7854 & \textbf{0.8149} & 0.8691 & 0.7887 & 0.7246 \\
  & noterm & 0.7971 &   -    & \textbf{0.8237} &   -    & 0.8107 &   -    \\
\midrule
\multirow{3}{*}{Qwen3\_SFT\_GRPO} 
  & proper & 0.7062 & \textbf{0.7053} & 0.7631 & \textbf{0.6208} & 0.7388 & \textbf{0.5545} \\
  & random & 0.6905 & \textbf{0.9577} & 0.7413 & \textbf{0.9525} & 0.7034 & \textbf{0.9656} \\
  & noterm & 0.7755 &   -    & 0.8118 &   -    & 0.7815 &   -    \\
  \midrule
\multirow{3}{*}{Gemma3\_SFT\_GRPO} 
  & proper & 0.7873 & 0.6980 & 0.8017 & 0.6097 & 0.8008 & 0.4669 \\
  & random & 0.7861 & 0.9106 & 0.8023 & 0.9337 & 0.7813 & 0.8351 \\
  & noterm & \textbf{0.8058} &   -    & 0.8023 &   -    & \textbf{0.8120} &   -    \\
\bottomrule
\end{tabular}
\caption{Results for language pairs: EN$\rightarrow$DE, EN$\rightarrow$ES, and EN$\rightarrow$RU. We report Comet-Kiwi and terminology match accuracy (denoted by term\%). Qwen3\_SFT and Gemma3\_SFT are the models after the first finetuning phase, and Qwen3\_SFT\_GRPO and Gemma3\_SFT\_GRPO the models after the GRPO phase. } 
\label{tab:results}
\end{table*}

Overall, we observe that baseline instruction-tuned models (Qwen3-4b, Gemma3-4b-it) achieve competitive COMET scores across all settings, with slightly higher scores in the \textit{no terminology} condition which is expected because the task is classical translation without any constraint. In terminology-constrained settings, \textit{proper terminology} leads to moderate term accuracy (around 0.40–0.55), whereas \textit{random terminology} consistently yields much higher term accuracy (0.77–0.96), as the random words from the input text are usually general meaning words and sometimes stop words.

After the first finetuning phase (Qwen3\_SFT, Gemma3\_SFT), term accuracy generally improves for both the \textit{proper} and \textit{random} terminology settings, while maintaining similar Comet-Kiwi scores. Finally, the GRPO phase produces a substantial boost in term accuracy for  \textit{proper} and \textit{random terminology} conditions. For example, Qwen3\_SFT\_GRPO achieves over 0.95 term accuracy in the \textit{random} setting across all language pairs. However, this improvement in terminology adherence comes at the cost of a moderate drop in COMET, particularly for Qwen3 models. Across models, Gemma3 variants maintain Comet-Kiwi more effectively after GRPO than Qwen3 variants, suggesting a better trade-off between translation quality and terminology accuracy. Therefore a good compromise model could be Gemma3\_SFT\_GRPO since it has  similar term accuracy except for the English-Russian pair but holds solid Comet-Kiwi scores. 

Language pair differences are also notable: EN$\rightarrow$ES generally attains slightly higher Comet-Kiwi scores than EN$\rightarrow$DE and EN$\rightarrow$RU, regardless of the setting, while EN$\rightarrow$RU tends to show the lowest term accuracy in \textit{proper terminology} for the base models (down to 0.40 for Gemma3-4b-it). This outcome is unsurprising, as the baseline systems already show weaker performance on EN$\rightarrow$RU. 

In summary, the proposed GRPO is highly effective when maximizing terminology adherence is the primary goal. However, if the aim is to balance translation quality and term accuracy, SFT-only models or Gemma3\_SFT\_GRPO provide a more favorable trade-off, as they retain higher Comet-Kiwi while still improving term accuracy.

\section{Conclusion}

In this work, we present our systems for the WMT25 Terminology Shared Task. We fine-tuned Qwen3-4b and Gemma3-4b-it in a two-stage process, combining supervised fine-tuning (SFT) with Group Relative Policy Optimization (GRPO). The fine-tuned models achieved substantial improvements in terminology match accuracy while maintaining strong COMET-Kiwi scores across all three evaluated language pairs. These results demonstrate the effectiveness of integrating terminology-aware training objectives into both SFT and reinforcement learning optimization, enabling models to better adhere to lexical constraints without sacrificing overall translation quality. Future work will explore extending terminology-aware training to additional language pairs and low-resource scenarios, investigating more sophisticated reward functions tailored for terminology adherence. Additionally, we plan to study the interaction between terminology control and other translation tasks such as inline tags to build an "all-in-one" multilingual and multi-task translation model.


\bibliography{custom}

\end{document}